\documentclass[11pt,a4paper]{article}
\usepackage[hyperref]{acl2018}
\usepackage{times}
\usepackage{latexsym}

\usepackage{url}
\usepackage{mathtools}
\usepackage{graphicx}
\graphicspath{{./figures/}}

\aclfinalcopy 

\title{Training a Neural Network in a Low-Resource Setting on Automatically Annotated Noisy Data}
	
\author{Michael A. Hedderich\textsuperscript{1,2} \hspace{1cm} Dietrich Klakow\textsuperscript{1} \\
\textsuperscript{1}Spoken Language Systems (LSV) \\
\textsuperscript{2}Saarbr\"ucken Graduate School of Computer Science \\
Saarland Informatics Campus, Saarland University, Saarbr\"ucken, Germany \\
{\tt \{mhedderich,dietrich.klakow\}@lsv.uni-saarland.de}
}

\date{}

\begin{document}
	\maketitle
	\begin{abstract}
		Manually labeled corpora are expensive to create and often not available for low-resource languages or domains. Automatic labeling approaches are an alternative way to obtain labeled data in a quicker and cheaper way. However, these labels often contain more errors which can deteriorate a classifier's performance when trained on this data. We propose a noise layer that is added to a neural network architecture. This allows modeling the noise and train on a combination of clean and noisy data. We show that in a low-resource NER task we can improve performance by up to 35\% by using additional, noisy data and handling the noise.
		
	\end{abstract}
		
	\section{Introduction}
	
	For training statistical models in a supervised way, labeled datasets are required. For many natural language processing tasks like part-of-speech tagging (POS) or named entity recognition (NER), every word in a corpus needs to be annotated. While the large effort of manual annotation is regularly done for English, for other languages this is often not the case. And even for English, the corpora are usually limited to certain domains like newspaper articles. For tasks in low-resource areas there tend to be no or only few labeled words available.
	
	Distant supervision and automatic labeling approaches are an alternative to manually creating labels. These exploit the fact that frequently large amounts of unannotated texts do exist in the targeted domain, e.g. from web crawls. The labels are then assigned using techniques like transferring information from high-resource languages \cite{das2011unsupervised} or simple look-ups in knowledge bases or gazetteers \cite{dembowski2017ner}. Once such an automatic labeling system is set up, the amount of text to annotate becomes nearly irrelevant, especially in comparison to manual annotation. Also, it is often rather easy to apply the system to different settings, e.g. by using a knowledge base in a different language.
	
	However, while easily obtainable in large amounts, the automatically annotated data usually contains more errors than the manually annotated. When training a machine learning algorithm on such noisy training data, this can result in a low performance. Furthermore, the combination of noisy and clean training instances can perform even worse than just using clean data.
	
	In this work, we present an approach to training a neural network with a combination of a small amount of clean data and a larger set of automatically annotated, noisy instances. We model the noise explicitly using a noise layer that is added to the network architecture. This allows us to directly optimize the network weights using standard techniques. After training, the noise layer is not needed anymore, removing any added complexity.
	
	This technique is applicable to different classification scenarios and in this work, we apply it to an NER task. To obtain a non-synthetic, realistic source of noise, we use look-ups from gazetteers for automatically annotating the data. In the low-resource setting, we show the performance boost obtained from training with both clean and noisy instances and from handling the noise in the data. We also compare to another recent neural network noise-handling approach and we give some more insight into the impact of using additional noisy data and into the learned noise model.
	
	\section{Related Work}
	\label{sec:related-work}
	
	Existing work showed the importance of handling label noise. \citet{Zhu2004ClassNV} suggested that noise in labels tends to be more harmful than noise in features. \citet{Beigman2009AnnotationNoise} showed that annotation noise in difficult instances can deteriorate the performance even on simple instances that would have been classified correctly in the absence of the hard cases.
	
	\citet{Rehbein2017Detecting} presented a technique for detecting annotation noise in automatically labeled POS and NER tags in an active learning scheme. It requires, however, several sources of automatic annotations and human supervision. Similarly, \citet{Rocio2007Detection} and \citet{Loftsson2009Correcting} focused on detecting noisy instances in (semi-) automatically annotated POS corpora, leaving the correction to human annotators.
	
	The model proposed by \citet{Bekker2016Unreliable} assumes that all clean labels pass through a noisy channel. One does only observe the noisy labels. The model of the noise channel, as well as the clean labels, are estimated using an EM algorithm. A neural network is then trained on the estimated labels. \citet{Berg2016NoisyLabel} applied this model to different tasks, obtaining small improvements on NER with automatically annotated data. A disadvantage of this approach is that the neural network needs to be retrained in every iteration of the EM algorithm, making the model difficult to scale to complex neural architectures.
	
	\citet{Goldberger2017NoiseAdaptation} transformed this model into an end-to-end trainable neural network by replacing the EM component with a noise adaptation layer. They experimented with simple image classification data and \citet{Dgani2018UnreliableMedical} applied it on the medical image domain. Both limit their approach to only using noisy data. Also, they just evaluate the effectiveness of their noise-handling method on simple synthetic noise (uniform and permutation). When applied to real-life scenarios, the noise might have a more complex structure.  
	
	In the image classification domain, several ideas have been proposed for estimating cleaned labels using a combination of clean and noisy labels. \citet{fergus2009semi} employ a label propagation approach. \citet{Sukhbaatar2015ConvolutionalNoisy} apply a noise model on top of a Convolutional Neural Network. \citet{vahdat2017toward} constructs an undirected graphical model to represent the relationship between clean and noisy labels. However, an additional source of auxiliary information is needed to infer clean from noisy labels. The approach presented by \citet{veit2017learning} uses two components. A cleaning network learns to map noisy labels to clean ones. The second network is used to learn the actual image classification task from clean and cleaned labels. We compare our approach to this idea in the experiments.
	
	\section{Noise Layer}
	\label{sec:noise-layer}
	
	Given a clean dataset $C$ consisting of feature and label tuples $(x,y)$, we can construct a multi-label neural network softmax classifier
	\begin{equation}
	p(y=i|x;w) = \frac{\exp(u_i^T h(x))}{\sum_{j=1}^k \exp(u_j^T h(x))}
	\end{equation}
	where $k$ is the number of classes, $h$ is a non-linear function or a more complex neural network and $w$ are the network weights including the softmax weights $u$.
	
	The noisy dataset $N$ is a set of additional training instances. Following the approach of \citet{Goldberger2017NoiseAdaptation}, we assume that each originally clean (but unseen) label $y$ went through a noise channel transforming it into the noisy label $z$. We only observe the noisy label, i.e. $N$ consists of tuples $(x,z)$.
	
	The noise transformation from a clean label $y$ with class $i$ to a noisy label $z$ with class $j$ is modeled using a stochastic matrix 
	\begin{equation}
	\theta(i,j) = p(z=j|y=i) = \frac{\exp(b_{ij}) }{\sum_{l=1}^k \exp(b_{il})}
	\end{equation}
	for $i,j \in \{1,...,k\}$ and where $b$ are learned weights. We call this the noise layer here. The probability for an observed, noisy label then becomes
	\begin{multline}
	p(z=j|x;w;\theta) = \\
	\sum^k_{i=1} p(z=j| y=i; \theta)p(y=i|x;w)
	\end{multline}
	for $(x,z) \in N$.
	
	\begin{figure}
		\includegraphics[width=\columnwidth]{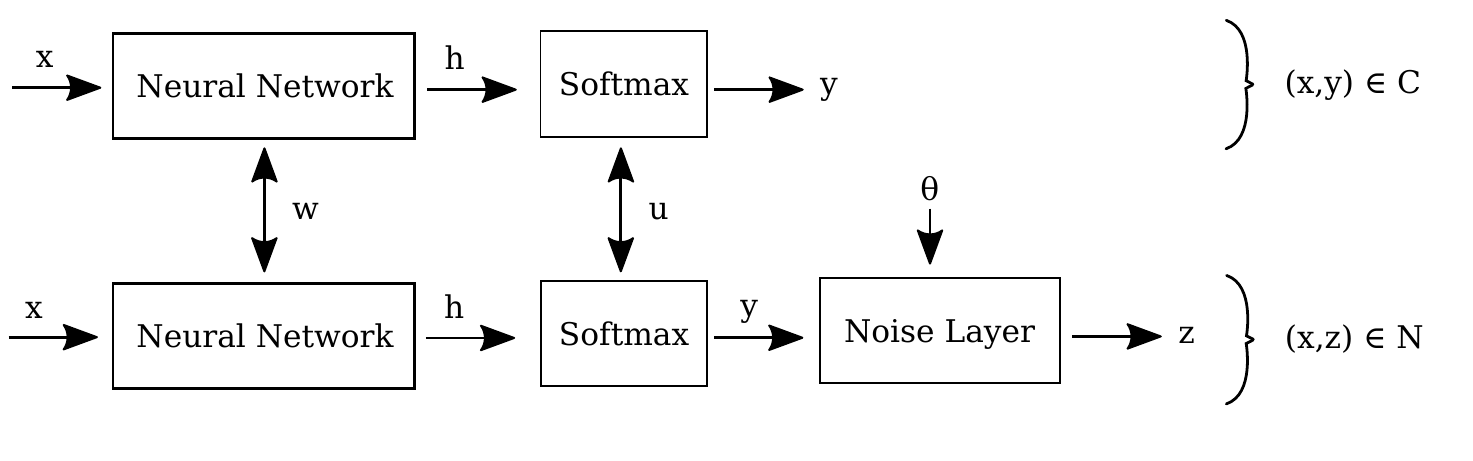}
		\caption{General architecture of the approach. Above is the base model trained on clean data $C$ and predicting clean labels $y$. Below is the noise layer model trained on noisy label data $N$. The predicted labels $y$ are transformed into the seen, noisy labels $z$ using the noise layer.}\label{fig:architecture}
	\end{figure}
	
	In contrast to the work by \citet{Goldberger2017NoiseAdaptation}, we also have access to clean data $C$. From this, we create two models, as illustrated in Figure \ref{fig:architecture}. The base model without noise layer is trained on $C$ and the noise model with the noise layer is trained on $N$. Both models share the same network weights. The models are trained alternatingly, each for one epoch of its corresponding clean or noisy data. For prediction, the noise layer is removed and just the base model is used.
	
	As stated by \citet{Goldberger2017NoiseAdaptation}, the initialization of $\theta$ is important. Since we have access to a small amount of clean data $C$, we use it for initializing the stochastic matrix. We assume that we can create noisy labels for the clean instances using the same process as for the noisy data $N$. We then initialize the weights of $\theta$ as 	
	\begin{equation}
	\label{eq:theta-init}
	b_{ij} = \log(\frac{\sum_{t=1}^{|C|} 1_{\{y_t=i\}} 1_{\{z_t=j\}}}{\sum_{t=1}^{|C|} 1_{\{y_t=i\}}}) 
	\end{equation} 	
	where $z_t$ is obtained by creating a noisy label for $(x_t,y_t) \in C$.
	
	\section{Dataset and Automatic Annotation}
	
	Named Entity Recognition (NER) is the task of assigning phrases in a text an entity label. In the sentence
	
	\begin{quote}
		\textit{Only France backed Fischler's proposal.}
	\end{quote}
	
	the country \textit{France} is of the entity class location and \textit{Fischler} refers to a person. Creating training data for this task requires that each word in the text is labeled with its corresponding class. The effort to create a sufficiently large dataset might be too large for a low-resource language. 
	
	To tackle this problem, \citet{dembowski2017ner} proposed to use external lists and gazetteers of entities to automatically label words in a training corpus. A list of person names can e.g be extracted from all of the entries appearing in Wikipedia's person category. Equipped with such lists for all entity classes, one can then label a text automatically. A word gets assigned a specific class if it appears in the corresponding entity list. A word or token that does not appear in any list gets assigned the null class "O". Additionally, simple heuristics help to resolve conflicts between lists and to remove some sources of errors. One might e.g. not label the day of the weeks as names, although \textit{"Friday"} might be in the list of person names.
	
	For this work, we use the English CoNLL-2003 NER corpus \cite{tjong2003conll}. The dataset is labeled with the classes person (PER), location (LOC), organization (ORG), miscellaneous name (MISC) and the null class (O). It consists of a training, a development and a test set. To obtain a low-resource setting, we randomly sample a subset of the training set as clean data $C$. In the experiments, we vary this size between ca. 400 and 20000 words. The rest of the labels are removed from the training set.
	
	We then label the whole training set using the method by \citet{dembowski2017ner} in the version with heuristics. This approach of automatically labeling words allows to quickly obtain large amounts of labeled text. However, both precision and recall tend to be lower than for manually labeled corpora (cf. Table \ref{tab:eval-autom-label}). It should be noted that the MISC class is not covered with this technique which is an additional source of noise in the automatically annotated data. We use this as our noisy data $N$.
	
	\begin{table}[t!]
		\begin{center}
			\begin{tabular}{|c|c|c|c|}
				\hline \bf Class & \bf Precision & \bf Recall & \bf F1 \\ \hline
				PER & 48.09\% & 25.90\% & 33.67\%  \\
				ORG & 52.45\% & 10.02\% & 16.83\%  \\
				LOC & 56.76\% & 65.42\% & 60.78\%  \\
				MISC & 0.00\% & 0.00\% & 0.00\% \\ \hline
				Overall & 53.31\% & 27.36\% & 36.16\% \\ \hline
			\end{tabular}
		\end{center}
		\caption{\label{tab:eval-autom-label}Evaluation of the automatic labeling on the full English CoNLL-2003 training set (which we use as noisy dataset $N$).}
	\end{table}
	
	\section{Model Architectures and Training}
	
	In this section, we present the different model architectures we evaluated in our experiments and we give details on the training procedure. 
	
	For each instance, the input $x$ is a sequence of words with the target word in the middle surrounded by 3 words from the left and from the right of the original sentence, e.g. x = \textit{"countries other than Britain until the scientific"} where \textit{"Britain"} is the target word with label $y = \text{LOC}$. Sentence boundaries are padded. We encode the words using the 300-dimensional GloVe vectors trained on cased text from Common Crawl \cite{pennington2014glove}. 
	
	The \textbf{base-model} uses a bidirectional LSTM \cite{hochreiter1997lstm} with state size 300 to encode the input. Then a dense layer is applied with size 100 and ReLU activation \cite{glorot2011relu}. Afterwards, the softmax layer is used for classification. This model is only trained on the clean data $C$.
	
	The \textbf{noise-model} is built upon the base model and uses the noise layer architecture explained in section \ref{sec:noise-layer}. First, the model is trained without noise layer for one epoch on the clean data. Then, we alternate between training with the noise layer on the noisy data and without the noise layer on the clean, each for one epoch. Instead of training on the full noisy corpus, we use a subsample $\tilde{N}$, randomly picked in each epoch. This allows the model to see many different noisy samples while preventing the noise from being too dominant. In section \ref{sec:exp-amount-noisy}, we evaluate this effect.
	
	For the \textbf{base-model-with-noise} we use the same clean and noisy data but the noise layer is left out, using only the base model architecture without an explicit noise-handling technique. 
	
	To evaluate the importance of the initialization of the stochastic matrix $\theta$, the \textbf{noise-model-with-identity-init} uses the same training approach and data as the \textit{noise-model}. However, $\theta$ is initialized with the identity matrix instead of using formula \ref{eq:theta-init}.
	
	The \textbf{noise-adaptation-model} uses the original model of \citet{Goldberger2017NoiseAdaptation}. It consists of the base model with the noise layer and is trained on the whole noisy dataset in each epoch. It does not use the clean data. For initializing $\theta$, the base model is pretrained on the noisy data and its predictions are used as an approximation to the clean labels.
	
	We also compare to the recent work by \citet{veit2017learning}. They train a noise cleaning component which learns to map from a noisy label and a feature representation to a clean label. These cleaned labels are then used for training of what we call the base model. The authors did not report specific layer sizes and their architecture is developed for an image classification task, which differs structurally from our NER dataset (e.g. their label vector is much sparser). We, therefore, adapt their concept to our setting. As feature representation, we use the output of the BiLSTM which is projected to a 30-dimensional space with a linear layer. This is concatenated with the noisy label and used as input to the noise cleaning component. It is passed through a dense layer with the same dimension as the label vector. The skip-connection and clipping are used as in their publication. We use the same training approach and data as with the \textit{noise-model}, replacing the step where the noise layer is trained. Instead, in each epoch the noise cleaning component is trained on $C$ and the corresponding noisy labels. The base model is then trained on a cleaned version of $\tilde{N}$ and $C$. We call this the \textbf{noise-cleaning-model}.
	
	All models are trained using cross-entropy loss, except for the noise cleaning component of the \textit{noise-cleaning-model} which is trained with the absolute error loss like in the original paper. All models are trained for 40 epochs and the weights of the best performing epoch are selected according to the F1 score on the development set. Adam \cite{kingma2015Adam} is used for stochastic optimization.
	
	\section{Experiments and Evaluation}
	
	In this section, we report on our experiments and their results. The training on noisy data as well as the randomness in training neural networks in general lead to a certain amount of variance in the evaluation scores. Therefore, we repeat all experiments five times and report the average as well as the standard error. To obtain meaningful results, no noise is added to the test data.
	
	\subsection{Model Comparison}
	
	To simulate different degrees of low-resource settings, we trained the models on different amounts of clean data. We vary the size between 407 labeled words (0.2\% of the CoNLL-2003 training data) and 20362 labeled words (10\%) in six steps. Since the noisy labels are easy to obtain, we use the whole corpus $N$. The size of the random subsample $\tilde{N}$ in each epoch is set to the same size as the clean data.
	
	\begin{figure}
		\includegraphics[width=\columnwidth]{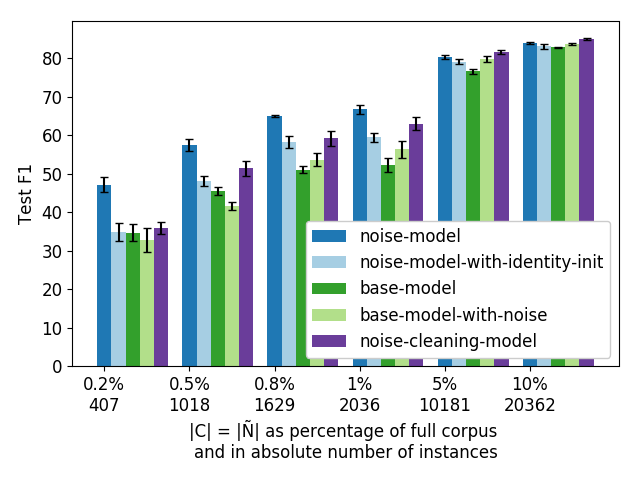}
		\caption{Evaluation results of the models. Experiments were run for different sizes of the clean data $C$ and the per epoch randomly subsampled data $\tilde{N}$. The average F1 score on the test set is given over five runs. The error bars show two-times standard error in both directions.}\label{fig:results-model-comparison} 
	\end{figure}
	
	The results of this experiment are given in Figure \ref{fig:results-model-comparison}. There is a general trend that the larger the amount of clean data is, the lower the differences between the models are. It seems that once we have obtained enough clean training data, the additional noisy data cannot add much more information, even when cleaned. This is reminiscent of results from semi-supervised learning (e.g. in \citealp{nigam2006semi}).
	
	For the two settings with the lowest amount of data, the \textit{base-model-with-noise} (which is trained on clean and noisy data without a noise channel) performs worst. For the four settings with more data, it is better than \textit{base-model} (which is only trained on $C$). This could indicate that noisy labels do hurt the performance in low-resource settings. However, once a certain amount of clean training data is obtained, this is enough to cope with the noise to a certain degree and obtain improvements, even when the noise is not explicitly handled.
	
	The models that do handle noise, outperform these baselines. When comparing \textit{noise-model} and \textit{noise-model-with-identity-init}, we see a large gap in performance. This shows the importance of a good initialization of the noise model $\theta$ in the low-resource setting.
	
	The original \textit{noise-adaptation-model} model by \citet{Goldberger2017NoiseAdaptation} obtains an average F1 score of 38.8. This shows that a model purely trained on a large amount of automatically annotated data can be an alternative to a model trained on very few clean instances. However, the effect of cleaning noisy labels without access to any clean data seems limited, as the model cannot even reach the performance of either the \textit{base-model} trained on 1018 instances nor our \textit{noise-model} on the smaller set of 407 instances.
	
	The \textit{noise-model} outperforms the \textit{cleaning-model} in the four lower-resource settings while the latter performs slightly better in the two scenarios with more data. With its access to the features in the noise cleaning component, the \textit{cleaning-model} might be able to model more complex noise transformations. However, it does not seem to be able to leverage this capability in a low-resource setting. In the low-resource settings, our \textit{noise-model} is able to handle the noise well and it gains over ten points in F1 score over not using a noise-handling mechanism or only training on clean data. 
	
	\subsection{Amount of Noisy Data}
	\label{sec:exp-amount-noisy}
	
	In this experiment, we evaluate the effect of using different amounts of noisy data during each epoch, i.e. we vary the size of the subsampled, noisy data $\tilde{N}$. We experiment with the \textit{noise-model} and fix the amount of clean data $C$ to 2036 labeled words (1\% of the CoNLL-2003 training data). We choose $|\tilde{N}|$ as multiples of $|C|$ using factors 0.5, 1, 2, 10, 20, 30 and 50. 
	
\begin{figure}
		\includegraphics[width=\columnwidth]{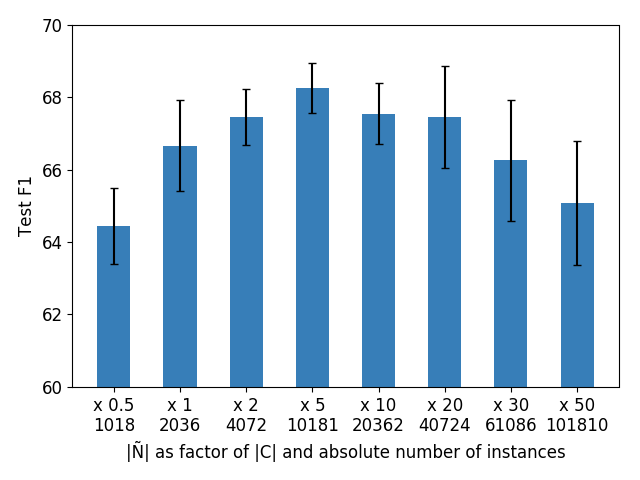}
		\caption{Evaluation results for varying the size of the per epoch randomly subsampled noisy data $\tilde{N}$. The \textit{noise-model} was used and the amount of clean data $C$ fixed to 2036 labeled words. The average F1 score on the test set is given over five runs. The error bars show two-times standard error in both directions.}\label{fig:results-amount-data-comparison} 
	\end{figure}

	The results are given in Figure \ref{fig:results-amount-data-comparison}. One can see a trend that increasing the size of $\tilde{N}$ results in an improvement in F1 score. This holds until factor 5. Afterwards, the performance degrades again. This might indicate that the noisy data becomes too dominant and the cleaning effect of the noise layer is not able to mitigate it.
	
	\subsection{Learned Weights}
	
\begin{figure}
		\includegraphics[width=\columnwidth]{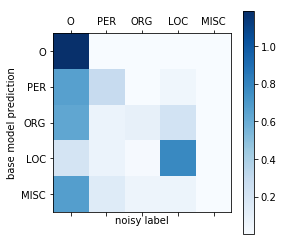}
		\caption{Representation of the noise transition weights $\theta$ learned in the noise layer. Each square is a value $\exp(\theta_{ij})$ where $i$ is the vertical and $j$ the horizontal index in the visualization. }\label{fig:results-learned-weights}
	\end{figure}

	Since, for evaluation purposes, we do have access to the clean labels of the whole training set, we can compare the noise that is in the noisy data to what the noise layer learned. Table \ref{tab:eval-autom-label} shows the evaluation of the automatically annotated labels on the training data. Figure \ref{fig:results-learned-weights} shows the stochastic matrix $\theta$ that was learned in one run of training the \textit{noise-model} with $|C| = |\tilde{N}| = $ 2036 labeled words (1\% of the CoNLL-2003 training data).
	
	One can see that the learned weight matrix represents a reasonable model of the noise. For the classes PER, ORG and MISC, the recall is very low in the noisy data and therefore the corresponding weights in the first column of the matrix are high: Instances (or a certain percentage of the probability mass) which the base model correctly classifies as PER/ORG/MISC, are mapped to the class O because this is the corresponding noisy label (indicated by the low recall). For the LOC class, the recall in the noisy labels is much higher and we see this reflected in the learned weights. The prominent weight is $\theta_{\text{LOC, LOC}}$, i.e. a prediction of the label LOC is mostly left unchanged because it tends to be correctly labeled in the noisy data.
	
	\section{Conclusions and Future Work}
	
	In this work, we presented a technique to train a neural network on a combination of clean and noisy annotations. We modeled the noise explicitly using a noise layer. We evaluated our approach on an NER task using real noise in the form of automatically annotated labels. We found that the probabilistic noise matrix learned is a useful model of the noise. In the low-resource setting where only a few manually annotated instances are available, we showed the improvements of up to 35\% obtained from using additional, noisy data and handling the noise. 
	
	For future work, we want to experiment with different classification tasks and other sources of noisy data. We would also like to explore more complex noise models that are able to perform well both in low- and high-resource settings.
	
	\bibliography{TrainingNoisyData}
	\bibliographystyle{acl_natbib}
	
	\appendix
	
\end{document}